\documentclass[runningheads]{llncs}
\usepackage{epsfig}
\usepackage{graphicx}
\usepackage{amsmath}
\usepackage{amssymb}
\usepackage{verbatim}
\usepackage{booktabs} 
\usepackage{adjustbox}
\usepackage{colortbl} 
\usepackage{xcolor} 
\usepackage{xfrac}
\usepackage[]{algorithm2e}
\usepackage{subcaption}
\usepackage{graphicx,multirow}
\usepackage{array}
\usepackage{colortbl}

\begin{document}
\title{Differential Morphed Face Detection Using Deep Siamese Networks\thanks{This work is based upon a work supported by the Center for Identification Technology Research and the National Science Foundation under Grant $\#1650474$.}}
%
%
\author{Sobhan Soleymani\inst{1}\inst{2}\inst{3}\orcidID{0000-0003-3541-0918} \and
Baaria Chaudhary\inst{1}\inst{2}\orcidID{0000-0002-9564-951X} \and
Ali Dabouei\inst{1}\orcidID{0000-0002-1084-6224} \and
Jeremy Dawson\inst{1}\orcidID{0000-0002-4539-7588} \and
Nasser M. Nasrabadi\inst{1}\orcidID{0000-0001-8730-627X}}
\authorrunning{S. Soleymani et al.}
%
\institute{West Virginia University, Morgantown WV 26506, USA\\ \and
{Authors Contributed Equally.}\\ \and Corresponding Author\\
\email{\{ssoleyma, bac0062, ad0046\}@mix.wvu.edu,\\ \{jeremy.dawson, nasser.nasrabadi\}@mail.wvu.edu}}
\maketitle              
\begin{abstract}
Although biometric facial recognition systems are fast becoming part of security applications, these systems are still vulnerable to morphing attacks, in which a facial reference image can be verified as two or more separate identities. In border control scenarios, a successful morphing attack allows two or more people to use the same passport to cross borders. In this paper, we propose a novel differential morph attack detection framework using a deep Siamese network. To the best of our knowledge, this is the first research work that makes use of a Siamese network architecture for morph attack detection. We compare our model with other classical and deep learning models using two distinct morph datasets, VISAPP17 and MorGAN. We explore the embedding space generated by the contrastive loss using three decision making frameworks using Euclidean distance, feature difference and a support vector machine classifier, and feature concatenation and a support vector machine classifier.

\end{abstract}

\section{Introduction}

Biometric facial recognition systems have increasingly been integrated into border control and other security applications that utilize identification tasks, such as official identity cards, surveillance, and law enforcement. These systems provide high accuracy at a low operational cost. In addition, face capture is non-invasive and benefits from a relatively high social acceptance. People use their faces to unlock their phones and also to recognize their friends and family. Furthermore, facial recognition systems contain an automatic fail-safe: if the algorithm triggers a false alarm, a human expert on-site can easily perform the verification. For these reasons, facial recognition systems enjoy a sizable advantage over other biometric systems. Consequently, the International Civil Aviation Organization (ICAO) has mandated the inclusion of a facial reference image in all electronic passports worldwide~\cite{icao20159303}. This means that the only biometric identifier present in passports globally is the face. 

Although facial recognition systems are largely successful, they still are not impervious to attack.  The mass adoption of automatic biometric systems in border control has revealed critical vulnerabilities in the border security scheme, namely the inability of these systems to accurately detect a falsified image. This vulnerability is further exacerbated by a loophole in the passport application process: the facial reference image, either digitally or as a physical print, is provided by the applicant at the time of enrollment.  This opens a window for the applicant to potentially manipulate the image before application submission. One type of manipulation that is recently identified as a serious threat is the morph attack~\cite{ferrara2014magic}, in which a facial reference image can be verified as two or more separate identities. A successful morphing attack allows two or more people to utilize the same passport to travel.

Thus, a criminal attacker, who otherwise cannot travel freely, could obtain a valid passport by morphing his face with that of an accomplice~\cite{ferrara2014magic}. Many morphing applications are not only freely available and easily accessible but also have no knowledge barrier~\cite{mallick2016face}. As such, it is almost absurdly simple for a criminal to procure a legitimate travel document. There are only a few straightforward steps: (1) find an accomplice with similar facial features, (2) morph both faces together such that existing facial recognition systems would classify the resulting morphed face as either of the original individuals, (3) the accomplice applies for a passport with the morphed image. The resulting passport could then be used by both the criminal and the accomplice to travel as they wish. Currently, we are unaware of any system in the passport verification process that is designed specifically for the detection of these manipulations. Moreover, commercial off-the-shelf systems (COTS) have repeatedly failed to detect morphed images~\cite{robertson2017fraudulent}. Likewise, studies show human recognizers are also unable to correctly differentiate a morphed facial image from an authentic one~\cite{robertson2017fraudulent,raghavendra2017face,bourlai2016face}.

\begin{table}[t]
\caption[Table caption text]{Table 1: Differential morph algorithms.}
\begin{center}
\begin{tabular}{ |c |c |c| }
\hline

Algorithm&Method&Database\\\hline
Face Demorphing&Image &MorphDB,\\
\cite{robertson2017fraudulent,Utrecht} &Subtraction &landmark-based\\\hline
Mutli-algorithm&Feature vectors and&Landmark-based \\
fusion approach~\cite{makrushin2017automatic}&feature difference& \\\hline

Deep models~\cite{damer2018morgan}&Feature embeddings&GAN-based\\
\hline
Our Method – Deep&$L_2$ difference of&landmarkbased,\\
Siamese Network&embedding representations& and GAN-based\\
\hline

\end{tabular}
\end{center}

\label{table:methods}
\end{table}

We propose to develop a novel differential morphing attack detection algorithm using a deep Siamese network. The Siamese network takes image pairs as inputs and yields a confidence score on the likelihood that the face images are from the same person. We employ a pre-trained Inception ResNET v1 as the base network. The experiments are conducted on two separate morphed image datasets: VISAPP17~\cite{makrushin2017automatic} and MorGAN~\cite{damer2018morgan,debiasi2019detection}. Results show an D-EER of 5.6\% for VISAPAP17 and an D-EER of 12.5\% for MorGAN. In the following sections, we briefly summarize the related works in Section II, explain the methodology in Section III, and discuss our experiments and subsequent results in Section IV. Finally, conclusions are presented in Section V.

\begin{figure}[t]
    \centering
    \includegraphics[width=\textwidth]{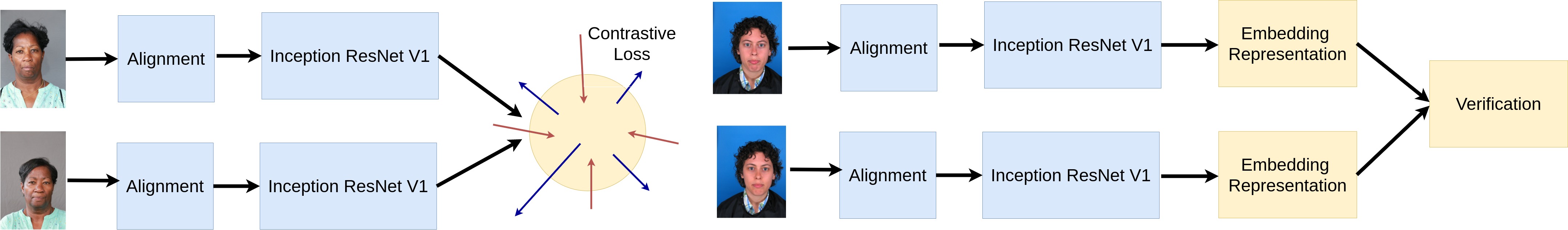}
    \caption{ Network architecture: Image pairs are fed into the MTCNN for face detection and alignment, then into the Inception ResNET v1,where contrastive loss is applied. From the feature embedding representation, verification is conducted by computing the $L_2$ distance between the feature vectors and a decision score is produced. Left) the training phase on WVU Twins Day dataset and the training portion of the morph dataset and right) the test phase. }
    \label{fig:archit}
\end{figure}

\section{Related Works}
The vulnerability of face recognition systems to morph attacks was first introduced by \cite{ferrara2014magic}. Since then, many morph detection algorithms have been proposed of two types: single (no reference) and differential. Single (no reference) morph attack detection algorithms rely solely on the potential morphed image to make their classification. Morphs are detected by extracting and analyzing features from the image in an attempt to identify the unique artifacts that indicate the face image was morphed or tampered with. On the other hand, differential morph attack detection algorithms rely on an additional trusted image, typically a live capture at border security, to compare the potential morphed image with. As such, differential morphing attack detection algorithms have more information at their disposal to make their classification and therefore perform significantly better than single morph detection algorithms \cite{scherhag2018towards}.

The majority of the current research exists solely in the single morph attack detection domain. Many classical hand-crafted feature extraction techniques have been explored to solve this problem. The most well-performing of these general image descriptors is Binarized Statistical Image Features (BSIF) \cite{kannala2012bsif}, used in \cite{raghavendra2016detecting}, in which extracted BSIF features were classified using a Support Vector Machine (SVM). Deep learning methods have also been proposed \cite{wandzik2017cnns} \cite{seibold2017detection} \cite{raghavendra2017transferable}. In \cite{raghavendra2017transferable}, complementary features from VGG-19 and AlexNet, both pre-trained on ImageNet and additionally fine-tuned on a morph dataset, are concatenated and then used to train a Probabilistic Collaborative Representation-based Classifier (ProCRC). A multi-algorithm fusion approach that combines texture descriptors, keypoint descriptors, gradient estimators, and deep neural network methods has also shown promising results \cite{scherhag2018morph}. The authors extract BSIF and Local Binary Pattern (LBP) \cite{liao2007learning} features to obtain feature vectors. Other feature descriptors such as Scale Invariant Feature Transform (SIFT) 
\cite{lowe2004sift} and Speeded-Up Robust Features (SURF) \cite{bay2006surf} are also used to extract keypoint descriptors and Histogram of Gradients (HOG) is used as a gradient estimator. Finally, deep feature embeddings from the OpenFace DNN are used as the last feature vector. All the above feature vectors are then used to train separate SVMs. In the end, score-level fusion is applied to obtain the final decision score for the potential morphed image.

There are a few papers also that address differential morph attack detection. Of these, reverting of a face morph or face demorphing \cite{ferrara2017face} \cite{ferrara2018face} has provided encouraging results. The demorphing algorithm subtracts the potential morphed image from the trusted image and uses the difference for classification. Feature extraction methods used in single morph attack detection can also be applied to the differential problem domain as well by taking the difference of the feature vectors of the potential morphed image and the trusted image. This difference vector along with the original feature vector for the potential morphed image were then used to train a difference SVM and a feature SVM, respectively. Score-level fusion is used to arrive at the final decision score. This method is explored in \cite{scherhag2018towards} using LBP, BSIF, SIFT, SURF, and HOG descriptors. Scherhag et. al \cite{scherhag2020deep} uses deep face representations from feature embeddings extracted from a deep neural network to detect a morph attack. The authors of \cite{scherhag2020deep} also emphasized the need for high-variance and constructed their morph image database using multiple morph algorithms. Several post-processing steps were also applied to emulate the actual compression methods used in storing passport photos in electronic passports, including reducing resolution, JPEG200 compression, and printing-and-scanning. Table~\ref{table:methods} compares the existing differential morph algorithms.

Although these methods have shown some success, none are sufficiently robust. These algorithms train on morph datasets of very limited size and scope. When tested on additional datasets, they perform poorly, indicating the models overfit \cite{scherhag2018performance}. These results are notably evident in the NIST FRVT morph detection test \cite{ngan2020face}, in which nearly all the algorithms submitted exhibited low performance on almost all tested morph datasets of varying quality and method. The NIST FRVT test is administered on multiple unseen datasets, ranging from automatically generated morphs to manually manipulated high quality morphs to print-and-scanned morphs. The deep learning method described in \cite{scherhag2020deep} outperforms the other models in the NIST test.

\section{Method}
The fundamental issue facing morph attack detection researchers is the lack of a large database of morphs with high variance. Many researchers create their own synthetic morph database, typically employing automated generative techniques, such as landmark manipulation \cite{mallick2016face} or General Adversarial Networks (GANs) \cite{damer2018morgan}. Although commercial software such as Adobe Photoshop or GIMP 2.10 have also been used to manually construct morphs, these methods are often time-consuming, and it is difficult to generate the number of morphs required to train a model. Each method generates different artifacts in the image, such as ghosting, unnatural transition between facial regions, hair and eyelash shadows, blurriness in the forehead and color, among others. 

As presented in Figure~\ref{fig:archit}, the proposed architecture is a Siamese neural network \cite{bromley1994signature}, in which the subnetworks are instances of the same network and share weights. Contrastive loss \cite{hadsell2006dimensionality}[25] is the loss function for training the Siamese network. Contrastive loss is a distance-based loss function, which attempts to bring similar images closer together into a common latent subspace, whereas it attempts to distance the dissimilar ones even more. Essentially, contrastive loss emphasizes the similarity between images of the same class and underscores the difference between images of different classes. The distance is found from the feature embeddings produced by the Siamese network. The margin is the distance threshold that regulates the extent to which pairs are separated.

\begin{equation}
L_c=(1-y_g)D(I_1,I_2)^2+y_g\max(0,m-D(I_1,I_2))^2
\label{eq:1}
\end{equation}

where $I_1$ and $I_2$ are the input face images, $m$ is the margin or threshold as described above and $y_g$ is the ground truth label for a given pair of training images and $D(I_1,I_2)$ is the L$_2$ distance between the feature vectors:

\begin{equation}
D(I_1,I_2)=||\phi(I_1)-\phi(I_2)||_2
\label{eq:2}
\end{equation}

Here, $\phi(.)$ represents a non-linear deep network mapping image into a vector representation in the embedding space. According to the loss function defined above, $y_g$ is $0$ for genuine image pairs and $y_g$ is $1$ for imposter (morph) pairs.

\begin{figure}
    \centering
    \includegraphics[width=300pt]{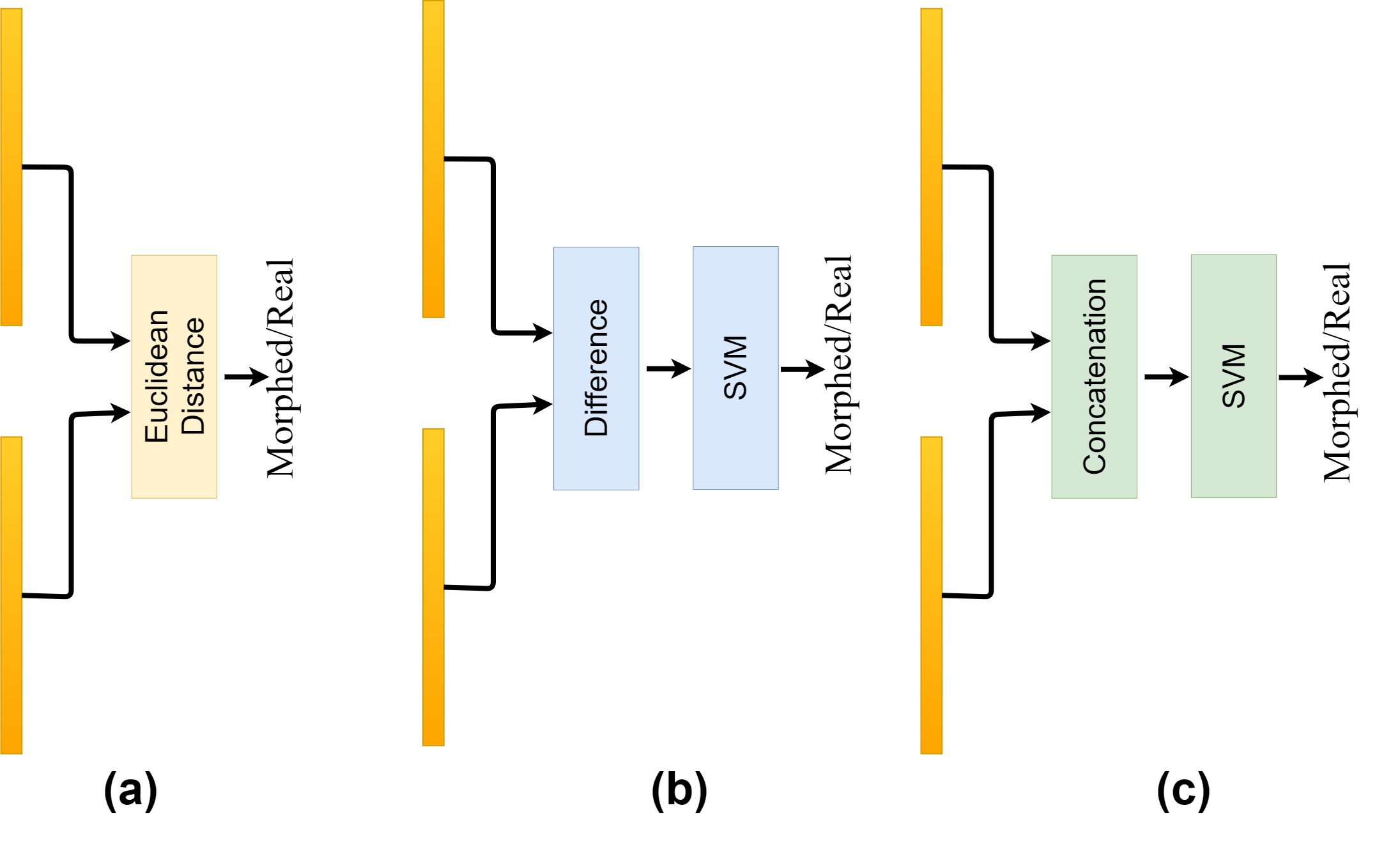}
    \caption{ During the test phase, three decision-making algorithms are considered to explore the embedding constructed embedding space: (a) Euclidean distance between the representation of the samples in a pair is considered to make the decision. (b) The difference and (c) the concatenation of the learned representations is fed to a SVM classifier. }
    \label{fig:svm}
\end{figure}

To streamline training, an Inception ResNET v1 architecture \cite{szegedy2017inception} is chosen as the base network, using weights pre-trained on the VGGFace2 dataset \cite{cao2018vggface2}. The network is then re-trained with the WVU Twins Day dataset [28] for the Siamese implementation. The model is optimized by enforcing contrastive loss on the embedding space representation of the genuine and imposter twin samples. The trained Siamese network is then additionally fine-tuned using the training portion of each morph database. To obtain a more discriminative embedding, the representations of the face image and its horizontal embedding are concatenated. The feature embeddings are taken from the last fully-connected layer and the $L_2$ distance between the two embeddings is calculated for the verification. As presented in Figure~\ref{fig:svm}, in our experiments we consider two additional decision making algorithms to explore the embedding space constructed by the contrastive loss, where we augment the proposed framework with the verification of the difference and concatenation of the embedding features of a pair using radial basis function kernel support vector machine (SVM) classifiers. These classifiers, which are learned using the training portion of each dataset, are utilized during the test phase to recognize genuine and imposter pairs.

\begin{table}[t]
\caption{The performance of the proposed framework on VISAPP17.}
\begin{center}
\addtolength{\tabcolsep}{-0pt}
\begin{tabular}{lccccccc}
\hline
\multirow{2}{*}{{Method}}&\multicolumn{3}{c}{{APCER@BPCER}}&\multicolumn{3}{c}{{BPCER@APCER}}&D-EER\\
&5\%&	10\%&	30\%	&5\%&	10\%&	30\%&\\
\hline

SIFT    & 45.12& 37.89& 17.94&       65.11& 43.28& 17.91     & 0.221\\
SURF    & 55.57& 42.72& 20.76&       72.58& 50.74& 20.89     & 0.225\\
LBP     &23.88 & 19.40& 1.58&        23.88& 20.65& 13.43     & 0.187\\
BSIF    & 25.37& 22.38& 1.49&        28.77& 25.37& 8.91      & 0.164\\
FaceNet & 11.82& 9.82 & 5.08&        29.82& 6.91 & 0.25      & 0.095\\
Ours    & 6.11 & 3.47 & 1.64&         7.31& 4.22 & 0.24      & 0.056\\
Ours+SVM&\multirow{2}{*}{5.78}&\multirow{2}{*}{3.29}&\multirow{2}{*}{1.52}&\multirow{2}{*}{6.67}&\multirow{2}{*}{3.95}&\multirow{2}{*}{0.21}&\multirow{2}{*}{0.054}\\
(concat.)\\
Ours+SVM&\multirow{2}{*}{5.29}&\multirow{2}{*}{3.17}&\multirow{2}{*}{1.43}&\multirow{2}{*}{6.12}&\multirow{2}{*}{3.71}&\multirow{2}{*}{0.19}&\multirow{2}{*}{0.052}\\
(difference)&\\
\bottomrule
\end{tabular}
\end{center}
\label{table:results_visapp}
\end{table}
\begin{table}[t]
\caption{The performance of the proposed framework on MorGAN}
\begin{center}
\addtolength{\tabcolsep}{-0pt}
\begin{tabular}{lccccccc}
\hline
\multirow{2}{*}{{Method}}&\multicolumn{3}{c}{{APCER@BPCER}}&\multicolumn{3}{c}{{BPCER@APCER}}&D-EER\\
&5\%&	10\%&	30\%	&5\%&	10\%&	30\%&\\
\hline
SIFT&	65.41&	53.37&	23.53&       	97.45&	66.66&	23.24&      	0.262\\
SURF&	69.88&	56.25&	29.82&	        98.24&	78.07&	30.06&	        0.298\\
LBP&	62.43&	54.13&	21.46&	        28.40&	18.71&	14.92&         	0.155\\
BSIF&	39.85&	31.26&	16.97&	        14.22&	 8.64&	 7.40&        	0.101\\
FaceNet&36.72&	30.15&	18.49&      	38.38&	26.67&	10.51&      	0.161\\
Ours&	31.85&	25.61&	13.21&      	14.32&	12.11&	5.49&       	0.125\\
Ours+SVM&\multirow{2}{*}{29.43}& \multirow{2}{*}{24.21}&  \multirow{2}{*}{12.35}&          \multirow{2}{*}{13.72}&  \multirow{2}{*}{11.75}&  \multirow{2}{*}{5.18}&           \multirow{2}{*}{0.113}\\ 
(concat.)&\\ 
Ours+SVM&\multirow{2}{*}{27.95}& \multirow{2}{*}{22.78}&  \multirow{2}{*}{12.05}&          \multirow{2}{*}{13.46}&  \multirow{2}{*}{10.42}&  \multirow{2}{*}{4.94}&           \multirow{2}{*}{0.102}\\
(difference)&\\ 
\bottomrule
\end{tabular}
\end{center}
\label{table:results_morgan}
\end{table}

\subsection{Experimental Setup}
\begin{figure}
    \centering
    \includegraphics[width=230pt]{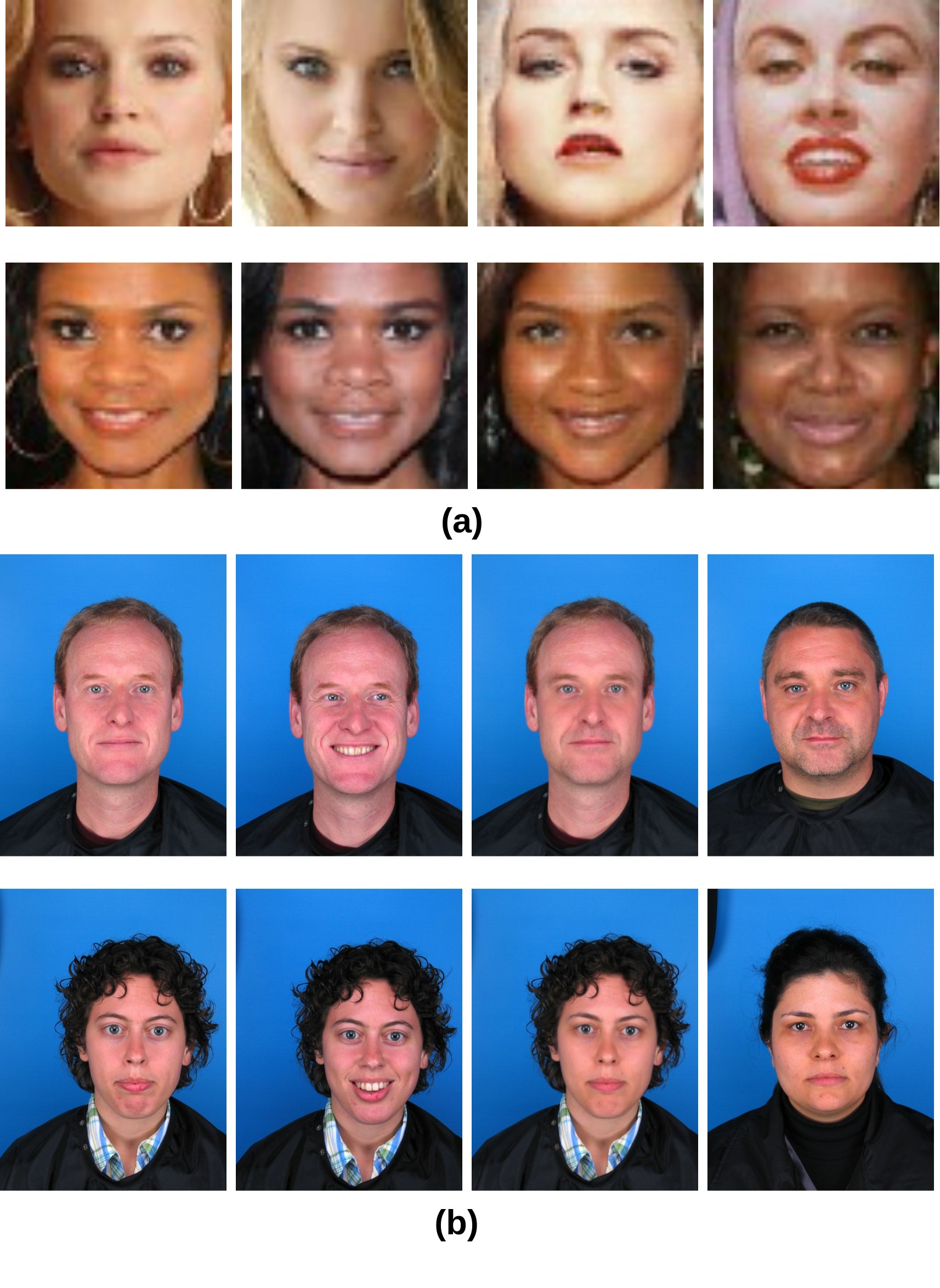}
    \caption{ Samples from (a) Morgan and (b) VISAPP17-Splicing-Selected datasets. For each dataset, the first and second faces are the gallery and probe bona fide images and the third face is the morph image construed from the first and forth face images. The original sizes for face images in these datasets are $64\times64$ and $1500\times 1200$, respectively. All the faces are resized to $160\times 160$ after detection and alignment using MTCNN.  }
    \label{fig:examples}
\end{figure}

The two morph image databases used in this experiment are VISAPP17 \cite{makrushin2017automatic} and MorGAN \cite{damer2018morgan} \cite{debiasi2019detection}. As presented in Figure~\ref{fig:examples}, we purposefully employ two different morph databases, created from two different face image datasets that apply two different morphing techniques to investigate how our model generalizes. VISAPP17 is a collection of complete and splicing morphs generated from the Utrecht FCVP database [30]. The images are $900\times1200$ pixels in size. This dataset is generated by warping and alpha-blending \cite{wolberg1998image}. To construct this dataset, facial landmarks are localized, the face image is tranquilized based on these landmarks, triangles are warped to some average position, and the resulting images are alpha-blended, where alpha is set to 0.5. A subset of 183 high quality splicing morphs constructed by selecting morph images without any recognizable artifacts (VISAPP17-Splicing-Selected dataset) is used along with 131 real images for a total set of 314 images. The morphs were created using the splicing technique, where after landmark manipulation, the resulting morphed face is spliced into the face of one of the original images. This preserves the background and hair, which helps avoid the issue of blurry artifacts and ghosting that typically occurs in these regions. However, this also means that the resulting morph derives its face shape from only one of the contributing individuals.

The MorGAN database is constructed from a selection of full-frontal face images manually chosen from the CelebA dataset \cite{liu2018large}. It consists of a custom morphing attack pipeline (MorGAN), created by the authors, that uses a GAN, inspired by inspired by learned inference model \cite{dumoulin2016adversarially}, to generate morphs. The database consists of 1500 bona fide probe images, 1500 bonafide references, and 1000 MorGAN morphs of $64\times 64$ pixels in size. There is also an additional MorGAN database, in which the MorGAN morphs have been super-resolved to $128\times 128$ pixels according to the protocol described in \cite{damer2019realistic}. The faces are detected and aligned using the MTCNN framework \cite{zhang2016joint}. The aligned images are then resized to $160\times 160$ pixels to prepare the images for the Siamese network. 

As the Siamese network expects pairs of images as input, the morph images are paired off into genuine face pairs and imposter face pairs, where a genuine pair consists of two trusted images and an imposter pair consists of a trusted image and a morph image. We employ the same train-test split provided by the authors of MorGAN to facilitate comparison of performance with other algorithms using this database. The train-test split is purposefully disjoint, with no overlapping morphs or contributing bonafides to morphs. This enables us to attain an accurate representation of the performance. For the VISAPP17 dataset, 50\% of the subjects are considered for training, while the other 50\% is used to evaluate the performance of the framework. For the train-test split we consider the same portions for male and female subjects. In addition, 20\% of the training set of the morph datasets was used during model optimization as the validation set. Batch size of 64 pairs of images of size $160\times160\times 3$ is used for training the model. Stochastic Gradient Descent (SGD) is the chosen optimizer. For the initial round of training with the Twins Day dataset, the initial learning rate is set to 0.1, multiplied by 0.9 every 5 epochs until the final value of $10^{-6}$. When fine-tuning with morph datasets, the initial learning rate is set to $10^{-3}$, then multiplied by 0.9 every 5 epochs until the final learning rate value of $10^{-6}$. The input data is further augmented with vertical and horizontal flips to increase the training set and improve generalization.

\section{Results}
We study the performance of the proposed differential morph detection framework using VISAPP17 and MorGAN datasets. The performance is compared with state-of-the-art classical and deep learning frameworks.

\subsection{Metrics} 
We apply the widely accepted metrics for morphing attack detection, APCER and BPCER, to our algorithm. The Attack Presentation Classification Error Rate (APCER) is the rate at which morph attack images are incorrectly classified as bonafide. Similarly, the Bonafide Probe Classification Error Rate (BPCER) is the rate at which bonafide images are incorrectly classified as morph attack presentations. In real-world applications, the BPCER is the measure by which individuals are inconvenienced with a false alarm. Hence, artificially regulating the BPCER rate by restricting it to fixed thresholds is recommended for face recognition systems \cite{frontex2012best}. We plot these rates in a Detection Error Tradeoff (DET) graph. D-EER is the detection Equal Error Rate or the decision threshold at which APCER and BPCER are equal. Additionally, we also present BPCER and APCER values for fixed APCER and BPCER values, respectively.

We employed BSIF \cite{kannala2012bsif} and LBP \cite{liao2007learning} as classical texture descriptors. In addition, we consider key-point detection frameworks, SURF \cite{bay2006surf} and SIFT \cite{lowe2004sift}. These classical methods have shown promise in morph detection in the literature. However, due to the private nature of the databases used in the original papers, we are unable to directly compare our results with theirs. Still, we employ the exact methodology described in the papers for best comparison. Our baseline models consist of these four frameworks in combination with an SVM classifier. The LBP feature descriptors are extracted according to patches of $3\times 3$. The resulting feature vectors, normalized histograms of size 256, are the values of the LBP binary code. The SIFT and SURF are implemented using the default parameters \cite{lowe2004sift} \cite{bay2006surf}. 8-bits BSIF feature vectors are constructed on a $3\times 3$ filters. These filters are Independent Component Analysis filters provided by \cite{hyvarinen2009natural}. The feature vectors are then fed into a SVM with an RBF kernel. For all classical baseline models, we follow \cite{scherhag2018towards}, where the feature representation of the image in question is subtracted from the feature representation of the trusted image before feeding it to the SVM classifier. 

We also compare our Siamese network with FaceNet \cite{schroff2015facenet} implementation Inception-ResNET v1, where the distance between the embedding representations of the images in pair is considered to provide the decision. Tables~\ref{table:results_visapp} and~\ref{table:results_morgan} presents the results on VISAPP17 and MorGAN datasets, respectively. As presented in these tables, fine-tuning using the training portion of each morph dataset demonstrably provides better performance for both datasets. This can be interpreted as the Siamese network is learning the generative nature of the morph images in the dataset. For the VISAPP17 train-test split, fine-tuning on the training portion of the dataset results in the BPCER@APCER=$5\%$ dropping from 29.82\% to 7.31\%. For the MorGAN dataset, this fine-tuning helps lower the BPCER@APCER=$5\%$ from 38.38\% to 14.32\%. The great difference in BPCER for MorGAN and VISAPP17 can be attributed to the overall difficulty of the MorGAN dataset, particularly because it is of a lower resolution than VISAPP17. On the other hand, the proposed SVM-based decision making frameworks can further improve the performance of the proposed framework.

For both datasets texture descriptors outperform key-point based models which is consistent with the study in [10]. In addition, the proposed Siamese framework outperforms the FaceNet implementation since the proposed framework initially learns to distinguish between the images with very small differences through re-training on WVU Twins Day dataset and then it is fine-tuned on the training portion of the corresponding dataset. For the MorGAN dataset, our results follow the original paper [8] results on single image morph detection, where the texture descriptors outperform convolutional neural networks, i.e., FaceNet.  This can be attributed to the generative nature of this dataset, which can be described using texture descriptors better than deep models. However, fine-tuning the deep model on the training portion of this dataset provides compatible results with BSIF.

\begin{figure}[t]
    \centering
    \includegraphics[width=300pt]{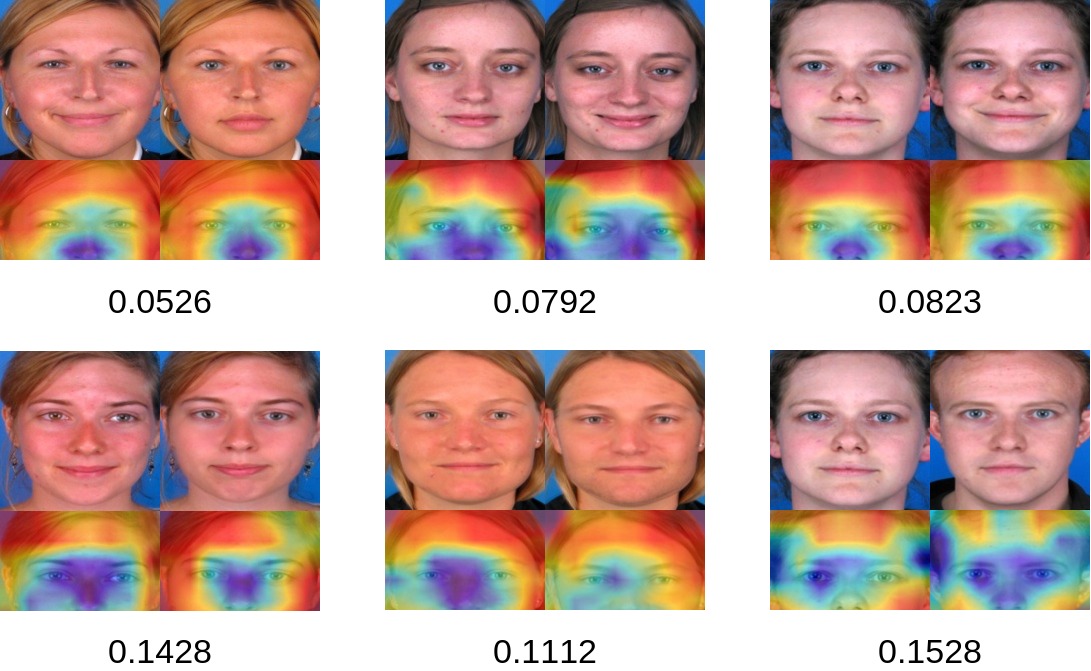}
    \caption{ Grad-CAMs for genuine (top) and imposter (bottom) pairs and the average per pixel distance between the images for each pair. For each imposter pair, the left and right images are real and morphed face images, respectively. }
    \label{fig:cam}
\end{figure}

Class activation maps~\cite{zhou2016learning} provide the attention of the decision with regard to regions of the face image. In Figure~\ref{fig:cam}, we follow the implementation of gradient-weighted class activation maps~\cite{selvaraju2017grad}. Here, we present the differentiation of the contrastive distance with regard to the feature maps constructed by ‘repeat\_2’ layer in Inception-ResNET v1. In this figure, we also report the average per pixel distance between the Grad-CAMs constructed for face images in each pair. As shown in these images, the difference between the activation maps for genuine pairs is smaller compared to the imposter pairs. It worth mentioning that since the datasets include neutral and smiley faces, while computing the distance between the activation maps, we do not consider the lower part of the faces. In addition, we can observe that the class activation maps for the two images in a genuine pair are roughly similar. 

\section{Conclusion}
In this paper, we proposed a deep Siamese network architecture to detect morphed faces. Using contrastive loss and a pre-trained Inception-ResNET v1 on WVU Twins Day dataset, we demonstrate the performance of our Siamese model on two different morph datasets. Likewise, we compare our model’s performance with baseline models constructed with common classical and deep methods employed in the literature, where our model outperforms the baseline models. This is attributed to the proposed framework learning to distinguish between images with small differences while training on WVU Twins Day dataset and learning the nature of the corresponding morph dataset by training on the training portion of the dataset.

\bibliographystyle{splncs04}
\bibliography{egbib}
\end{document}